\documentclass[conference]{IEEEtran}
\IEEEoverridecommandlockouts

\usepackage{cite}
\usepackage{amsmath,amssymb,amsfonts}
\usepackage{algorithmic}
\usepackage{graphicx}
\usepackage{textcomp}
\usepackage{xcolor}
\usepackage{hyperref}

\def\BibTeX{{\rm B\kern-.05em{\sc i\kern-.025em b}\kern-.08em
    T\kern-.1667em\lower.7ex\hbox{E}\kern-.125emX}}
\begin{document}

\title{ChronoPscychosis: Temporal Segmentation and Its Impact on Schizophrenia Classification Using Motor Activity Data\\

}

\author{\IEEEauthorblockN{ Pradnya Rajendra Jadhav}
\IEEEauthorblockA{\textit{Department of Biological Sciences and E\&I} \\
\textit{BITS Pilani, K.K. Birla Goa campus}\\
Goa, India \\
f20191135@goa.bits-pilani.ac.in}
\and
\IEEEauthorblockN{Raviprasad Aduri}
\IEEEauthorblockA{\textit{Department of Biological Sciences} \\
\textit{BITS Pilani, K.K. Birla Goa campus}\\
Goa, India \\
aduri@goa.bits-pilani.ac.in}

 }
\maketitle

\begin{abstract}
Schizophrenia is a complicated mental illness characterized by a broad spectrum of symptoms affecting cognition, behavior, and emotion. The task of identifying reliable biomarkers to classify Schizophrenia accurately continues to be a challenge in the field of psychiatry. We investigate the temporal patterns within the motor activity data as a potential key to enhancing the categorization of individuals with Schizophrenia, using the dataset having motor activity recordings of 22 Schizophrenia patients and 32 control subjects. The dataset contains per-minute motor activity measurements collected for an average of 12.7 days in a row for each participant. We dissect each day into segments (Twelve, Eight, six, four, three, and two parts) and evaluate their impact on classification. We employ sixteen statistical features within these temporal segments and train them on Seven machine learning models to get deeper insights. LightGBM model outperforms the other six models. Our results indicate that the temporal segmentation significantly improves the classification, with AUC-ROC = 0.93, F1 score = 0.84( LightGBM- without any segmentation) and AUC-ROC = 0.98, F1 score = 0.93( LightGBM- with segmentation). Distinguishing between diurnal and nocturnal segments amplifies the differences between Schizophrenia patients and controls. However, further subdivisions into smaller time segments do not affect the AUC-ROC significantly. Morning, afternoon, evening, and night partitioning gives similar classification performance to day-night partitioning. These findings are valuable as they indicate that extensive temporal classification beyond distinguishing between day and night does not yield substantial results, offering an efficient approach for further classification, early diagnosis, and monitoring of Schizophrenia.
\end{abstract}

\vspace{5pt} 


\section{\textbf{Introduction}}
Mental illnesses deeply impact an individual's thoughts, perceptions, and self-awareness \cite{b1}. Schizophrenia is one of the severe mental disorders that substantially burden patients and society. It exhibits positive symptoms like hallucinations (often auditory), delusions such as persecution and passivity, and disrupted thinking. Schizophrenia also involves negative symptoms, including social withdrawal, reduced motivation, and emotional numbness \cite{b2}. Schizophrenia is a multifaceted disorder stemming from genetics, environment, or their interplay. Variables such as birth complications, childhood trauma, substance abuse, and migration have been identified as potential risk factors contributing to the development of this condition \cite{b3}. 
\vspace{5pt} 

Disruption of the circadian and sleep cycles are commonly seen in people with schizophrenia. These disturbances affect approximately 80\% of patients with the disorder \cite{b4}. Patients frequently experience disturbed sleep patterns, especially insomnia, even prior to the start of schizophrenia. These patients experience insomnia due to a variety of circumstances, including a lack of regularity and activity during the day, an obsession with commencing and staying asleep, and the emergence of uncomfortable thoughts or hallucinations when trying to fall asleep \cite{b5}. Given the significance of day and night activities as crucial classifiers in distinguishing between Schizophrenia patients and controls, the implementation of temporal segmentation can provide valuable insights for the early diagnosis and monitoring of Schizophrenia.
\vspace{5pt} 

American Psychiatric Association's DSM-5 manual is the often utilized for diagnosing Schizophrenia \cite{b6}. This method is time-consuming as it requires frequent consultations from mental health professionals. Alternatively, extensive research has been conducted using EEG signal analysis with machine learning and deep learning models for schizophrenia diagnosis \cite{b7},\cite{b8}. EEG signals are recorded by positioning electrodes on a person's scalp to measure brain activity \cite{b9}. However, acquiring these measurements over an extended period, particularly at night, can be challenging. Collecting motor activity data (Actigraphy data) through smartwatches that have motion sensors is feasible over an extended time, which makes further temporal segmentation easier.
\vspace{5pt} 

Actigraph devices track variables like overall sleep time, interrupted sleep, abrupt awakenings, and efficiency of sleep. These measures have been used to identify various health-related outcomes \cite{b10}. This actigraphy data has been utilized over the past decade to explore different diagnosis methods for mental illnesses like depression, Parkinson's disease, and Alzheimer's \cite{b11}.

\vspace{5pt} 
The treatment and diagnosis of mental disorders has traditionally incorporated an examination of sleep-wake patterns \cite{b12}. Studies have also explored irregularities within the 24-hour circadian rhythms as potential diagnostic markers for mental illnesses \cite{b13}, \cite{b14}. This study primarily focuses on using motor activity data acquired via smartwatches to implement temporal segmentation and determine its effectiveness. This study aims to establish an optimized data segmentation approach, eliminating inefficiencies associated with improper segmentation.

\section{\textbf{Related work}}
The dataset used in this research was created by Jakobsen et al.\cite{b15} They analyzed this dataset with three features and without any segmentation. Fellipe et al. created nine statistical features and segmented the day into three parts for this dataset. They tested this dataset on Five machine learning models along with one deep learning model mainly to improve accuracy and precision \cite{b16}. Boeker et al. used Hidden Markov model specifications \cite{b17}. Misgar et al. used UMAP features for the analysis of this dataset \cite{b18}. Sheha et al. worked on the feature engineering of the dataset of patients with depression and Schizophrenia. They worked on daily and quarter-daily segments and suggested a deeper analysis of shorter time intervals for future work \cite{b19}. As far as we are aware, the study highlighting comparative analysis of 8 different time segmentation patterns has not been performed in the scenario of Schizophrenia. We conducted a comparative study to optimize and identify one best-performing time interval.

\section{\textbf{Dataset details}}

This research uses motor activity data collected through the wristwatches called Actiwatch (model AW4) developed by Cambridge Neurotechnology Ltd, England. This device contains a particular sensor (piezoelectric accelerometer) that tracks movement intensity, amount, and duration in different directions (x, y, and z axes). This device recorded movements with a frequency of 32 Hz. It generated an integer value proportional to the movement's intensity in a one-minute interval \cite{b15}.

\vspace{5pt} 

The dataset has actigraphy data of 22 individuals diagnosed with Schizophrenia. These individuals were under long-term care in a psychiatric ward at Haukeland University Hospital. Among these patients, 19 were males and 3 were females, with an average age of 46.9 years. Their initial hospitalization occurred at an average age of 24.8 years. The diagnosis of these patients was conducted by psychiatrists using the DSM-IV manual. Specifically, 17 patients were diagnosed with paranoid Schizophrenia, while the subtype for 5 patients was not specified. The current DSM-5 manual does not recognize schizophrenia subtypes.

\vspace{5pt} 

The dataset includes data from 32 healthy individuals who served as a control group. None of them had a history of any psychiatric disorder. There were 20 females and 12 males in this group, with 38.2 years as the mean age. The patients and the control group used the actigraphy device for an average of 12.7 days.
\vspace{5pt} 

The PSYKOSE dataset can be accessed through the following link: https://osf.io/dgjzu/ 

\begin{figure}[htbp]
   \centering
   \includegraphics[width=1\linewidth]{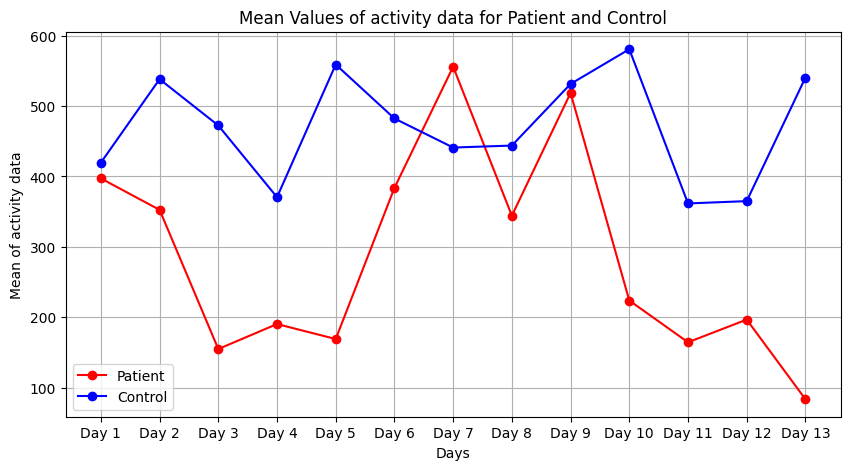}
   \caption{Hourly daytime mean of motor activity for a patient and control. There is no consistent pattern in the mean activities hence the temporal segmentation is needed. }
    \label{Figure:first-label}
 \end{figure}

Figure. 1 displays the daytime mean activity levels of a patient and control for 13 days. There is a variation in the pattern of mean activity, with some days showing higher mean activity for patients while controls exhibit higher activity levels on other days. This fluctuation presents a challenge in the classification process. Temporal segmentation plays a critical role in identifying distinct patterns between the control group and the group of schizophrenia patients, improving the overall accuracy of the classification process.

\section{\textbf{Data segmentation and modeling}}

\subsection{Data cleaning and processing}\label{AA}
Prioritizing temporal segmentation was a critical aspect. Only the data available for a full day (24 hours) was considered to assure the temporal pattern's integrity. If the data was only available for a few hours on any particular day, it was discarded. The dataset was structured into distinct columns, namely date, timestamp, and class, where a class value of "1" represented patients, and "0" indicated control subjects. This column organization helped develop a more streamlined segmentation process.

\subsection{Segmentation of the data}

As shown in Figure. 1, analyzing the data as a whole was not fruitful for classification. Therefore, the data was divided into eight distinct patterns to evaluate classification efficiency.

\vspace{-\baselineskip}
\begin{table}[htbp]
\caption{Temporal Patterns and Time Intervals}
\begin{center}
 \begin{tabular}{|p{3cm}|p{5cm}|}
 \hline
 \multicolumn{1}{|c|}{\textbf{Temporal Pattern}} & \multicolumn{1}{c|}{\textbf{Time Intervals}} \\

 \hline
 24 hours in 2 Parts & 08:00 - 19:59 and 20:00 - 07:59  \\
 \hline
 24 hours in 3 Parts & 00:00 - 07:59, 08:00 - 15:59, 16:00 - 23:59 \\
 \hline
 24 hours in 4 Parts & 00:00 - 05:59, 06:00 - 11:59, 12:00 - 17:59, 18:00 - 23:59 \\
 \hline
 24 hours in 6 Parts & 00:00 - 03:59, 04:00 - 07:59, 08:00 - 11:59, 12:00 - 15:59, 16:00 - 19:59, 20:00 - 23:59 \\
 \hline
 24 hours in 8 Parts & 00:00 - 02:59, 03:00 - 05:59, 06:00 - 08:59, 09:00 - 11:59, 12:00 - 14:59, 15:00 - 17:59, 18:00-20:59, 21:00-23:59 \\
 \hline
 24 hours in 12 Parts & 00:00 - 01:59, 02:00 - 03:59, 02:00 - 03:59, 04:00 - 05:59, 06:00 - 19:59, 08:00 - 09:59, 10:00 - 11:59, 12:00 - 13:59, 14:00 - 15:59, 16:00 - 17:59, 18:00 - 19:59, 20:00 - 21:59, 22:00 - 23:59 \\
 
 \hline
 
 24 hours (Full 1 day) & 00:00 - 24:00   \\
 \hline
 All days together   & 12.7 days on average  \\
 \hline
 \end{tabular}
 \label{tab1}
 \end{center}
 \end{table}
\vspace{20pt}
Table I presents the temporal patterns used in our analysis. Initially, we considered the entire 24-hour day. Subsequently, we systematically divided this time span into 2, 3, 4, 6, 8 and 12 segments to explore the importance of temporal segmentation in the classification process.
The segmentation methodology involved observation of activity data within specific time intervals. 
\vspace{5pt}

In the case of the two-part segmentation, we based our division on the intensity of motor activity observed in both patients and control subjects. The period from 08:00 to 20:00 exhibited high activity levels and was classified as "day." On the other hand, the period from 20:00 to 08:00 was categorized as "night" due to reduced activity associated with sleep. This night segment was particularly crucial for identifying disruptions in patients' sleep patterns. Furthermore, to determine the time intervals for the 3, 4, 6, 8 and 12-part segmentation, we precisely analyzed the patterns in the differences in the motor activity data of the patient and control group and symmetrically divided the time intervals.
\vspace{10pt}

\vspace{-\baselineskip}
\subsection{Feature extraction}
Statistical features play a crucial role in quantitatively describing, comparing, and analyzing the motor activity data \cite{b20}. They contribute to pattern recognition and dimensionality reduction, which consequently helps construct models and train the data.
\begin{figure}[htbp]
   \centering
   \includegraphics[width=1\linewidth]{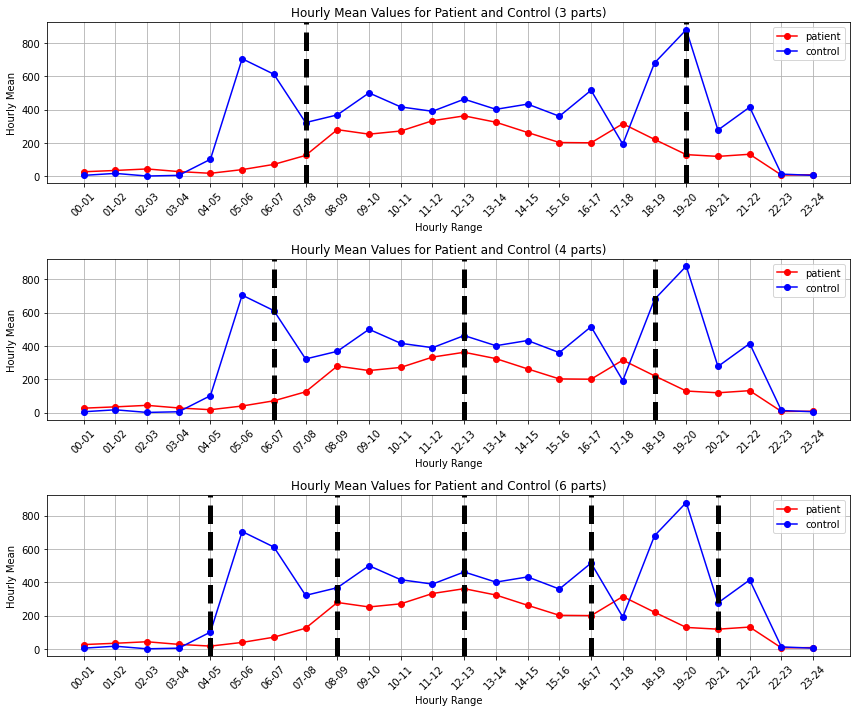}
   \caption{Segmentation of 24 hours in 3,4 and 6 parts. Segmenting data into distinct parts enhances classification efficiency by recognizing and utilizing the unique patterns within each segment, rather than treating the data as a unified whole. }
    \label{fig:first-label}
 \end{figure}

Figure. 2 represents the segmentation of a day in 3,4, and 6 intervals. The minimum values of the data show striking similarities between patient and control groups within each segment. This doesn't contribute to the classification, therefore it is discarded. However, certain statistical features, including mean, median, and maximum value, show variations between patients and controls in these segments. These are included in the analysis.

\vspace{5pt}

Based on the data distribution, we used the following sixteen statistical features for the classification- Mean, Median, Standard Deviation (Std Dev), Proportion of Zeros(Indicates the percentage of zero values in the data), Skewness(Describes the asymmetry of the data distribution), Kurtosis(Gives insights about the shape of the data distribution), Maximum (Max), Median Absolute Deviation(MAD-Quantifies the variability of the data around the median), Interquartile Range(IQR-Range of the middle data values), Coefficient of Variation(CV- Measures how much the data fluctuates in comparison to the average), Entropy (Measures degree of unpredictability), Autocorrelation (Assesses temporal patterns and dependencies), Number of peaks and troughs, Semivariance, Root mean square (Offers insights into the overall strength of motor activity over time).
\vspace{5pt}

The importance of features for the morning and night periods, calculated using LightGBM's gain metric is presented in Figure. 3. The most crucial feature is 'Night- Autocorrelation.' This feature has the highest impact on the model's prediction. Night proportion of zeros and standard deviation also has high significance. This is understandable since it signifies the presence of sleep disturbances in patients, distinguishing them from individuals in the control group. Morning autocorrelation and standard deviation also contribute significantly to the prediction for the model, primarily due to the reduced motivation and diminished movement observed in schizophrenia patients in the morning periods.

 \begin{figure}[htbp]
   \centering
   \includegraphics[width=1\linewidth]{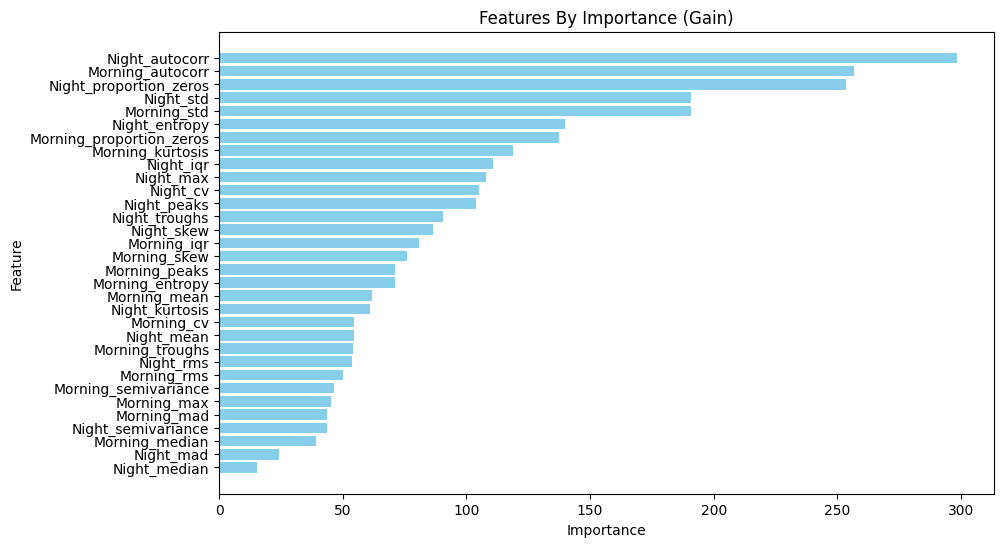}
   \caption{Importance of features in prediction of the model (LightGBM- 10 cross fold)}
    \label{fig:first-label}
 \end{figure}

\subsection{Machine Learning Modeling and Evaluation Metrics}

Selecting machine learning models and metrics is of utmost importance for ensuring accurate classification and clinical relevance. The models discern patterns, relationships, and distinctions essential for classification by processing and analyzing the statistical features. Initially, we had a set of ten statistical features to encapsulate the characteristics of a 24-hour day. Upon temporal segmentation into 2, 3, 4, 6, 8 and 12 parts, the feature count correspondingly expanded to 32, 48, 64, 96, 128 and 192, respectively.

\vspace{5pt}
In our study, we employed seven machine learning models. LightGBM \cite{b21} and XGBoost \cite{b22} are known for achieving high predictive accuracy, Random Forest for its overfitting prevention \cite{b23}, Logistic Regression for its interpretability \cite{b24}, Support Vector Machine, K- nearest neighbours, Decision Trees for their capacity to provide clear insights. We evaluated these models by applying 10-fold cross-validation.

\vspace{5pt}
As our primary goal is to investigate temporal segmentation, we have strongly emphasized two key metrics: F1-score along with the AUC-ROC (Area Under the Receiver Operating Characteristic Curve). AUC-ROC is critical for measuring the accuracy of our models, enabling us to identify their ability to distinguish between schizophrenia patients and control subjects \cite{b25}. On the other hand, the F1 score plays a crucial role in striking a balance between precision and recall \cite{b26}. These metrics help us achieve both high accuracy and a well-rounded evaluation of our models' performance as we study the complexities of temporal segmentation.

\subsection{Hardware and Softwares utilized }

In this study, the computations were done using MacBook Air(2017) with macOS Monterey version 12.4. The hardware configuration of the system included 1.8 GHz Dual-Core Intel Core i5 processor, 8 GB RAM and 128 GB SSD storage. The programming language used was Python3 and the code was executed through VScode. Jupyter notebook was utilized to generate the figures.

\section{\textbf{Results}}

After the feature engineering and model selection, we generated CSV files for various temporal segmentation scenarios. These scenarios included a full 24-hour day, divisions into 2, 3, 4, 6, 8 and 12 segments within a 24-hour period, and a final scenario with all days of observation (an average of 12.7 days for each patient/control) combined into a single sheet.


\begin{table}[htbp]
\caption{AUC ROC and F1 Scores for Different Temporal Segmentation Patterns and Machine Learning Models}
\begin{center}
\begin{tabular}{|p{2cm}|p{2.5cm}|p{1cm}|p{1cm}|}
\hline
\textbf{Temporal Segmentation} & \textbf{Model} & \textbf{AUC ROC} & \textbf{F1 score} \\
\hline
12 Parts & Lightgbm & 0.98 & 0.90 \\
& XGboost & 0.97 & 0.89 \\
& Random Forest & 0.96 & 0.88 \\
& Logistic Regression & 0.94 & 0.85 \\
& SVM & 0.95 & 0.86 \\
& K-Nearest Neighbors & 0.92 & 0.84 \\
& Decision Trees & 0.81 & 0.77 \\
\hline
8 Parts & Lightgbm & 0.97 & 0.90 \\
& XGboost & 0.98 & 0.89 \\
& Random Forest & 0.96 & 0.87 \\
& Logistic Regression & 0.93 & 0.84 \\
& SVM & 0.93 & 0.83 \\
& K-Nearest Neighbors & 0.90 & 0.83 \\
& Decision Trees & 0.83 & 0.81 \\
\hline
6 Parts & Lightgbm & 0.98 & 0.90 \\
& XGboost & 0.97 & 0.89 \\
& Random Forest & 0.96 & 0.87 \\
& Logistic Regression & 0.92 & 0.83 \\
& SVM & 0.93 & 0.83 \\
& K-Nearest Neighbors & 0.90 & 0.82 \\
& Decision Trees & 0.82 & 0.79 \\
\hline
4 Parts & Lightgbm & 0.97 & 0.89 \\
& XGboost & 0.97 & 0.89 \\
& Random Forest & 0.96 & 0.86 \\
& Logistic Regression & 0.90 & 0.80 \\
& SVM & 0.92 & 0.81 \\
& K-Nearest Neighbors & 0.89 & 0.79 \\
& Decision Trees & 0.84 & 0.81 \\
\hline
3 Parts & Lightgbm & 0.97 & 0.90 \\
& XGboost & 0.97 & 0.88 \\
& Random Forest & 0.97 & 0.87 \\
& Logistic Regression & 0.92 & 0.83 \\
& SVM & 0.92 & 0.83 \\
& K-Nearest Neighbors & 0.88 & 0.79 \\
& Decision Trees & 0.86 & 0.83 \\
\hline
2 Parts & Lightgbm & 0.97 & 0.90 \\
& XGboost & 0.97 & 0.88 \\
& Random Forest & 0.97 & 0.88 \\
& Logistic Regression & 0.92 & 0.83 \\
& SVM & 0.92 & 0.83 \\
& K-Nearest Neighbors & 0.88 & 0.79 \\
& Decision Trees & 0.86 & 0.84 \\
\hline
Full Day & Lightgbm & 0.95 & 0.91 \\
& XGboost & 0.95 & 0.92 \\
& Random Forest & 0.95 & 0.89 \\
& Logistic Regression & 0.91 & 0.86 \\
& SVM & 0.89 & 0.85 \\
& K-Nearest Neighbors & 0.85 & 0.84 \\
& Decision Trees & 0.82 & 0.85 \\
\hline
All Days & Lightgbm & 0.93 & 0.84 \\
& XGboost & 0.92 & 0.90 \\
& Random Forest & 0.92 & 0.91 \\
& Logistic Regression & 0.87 & 0.77 \\
& SVM & 0.91 & 0.85 \\
& K-Nearest Neighbors & 0.90 & 0.90 \\
& Decision Trees & 0.83 & 0.85 \\
\hline
\end{tabular}
\label{tab_aucroc_f1}
\end{center}
\end{table}

\subsection{Performance of ML Models}
Table II displays the AUC ROC and F1 score values resulting from 10-fold cross-validation across various segmentation patterns for the five machine learning models. LightGBM model outperforms the other six ML models in most of the segmentation patterns, consistently yielding higher AUC-ROC and F1 scores. Following LightGBM, XGBoost, and Random Forest exhibit closely competitive performance across most scenarios.

\vspace{5pt}
The highest AUC-ROC values were achieved by the LightGBM model when the 24-hour segmentation was divided into 6 and 12 parts.

\subsection{Significance of Temporal Segmentation}

Figure. 4 presents the AUC-ROC values for different temporal patterns. The division of the 24 hour period into segments significantly enhanced the AUC-ROC values. While considering the entire 24-hour cycle, the AUC-ROC was 0.95 in LightGBM model. However, after segmentation into 2, 3, 4, 8 parts, it increased to 0.97. When segmentation was done with 6 and 12 parts, this value was slightly increased to 0.98 but the F1 score is very similar in all 2, 3, 4, 6, 8 and 12 parts segmentation patterns.

\begin{figure}[htbp]
   \centering
   \includegraphics[width=1\linewidth]{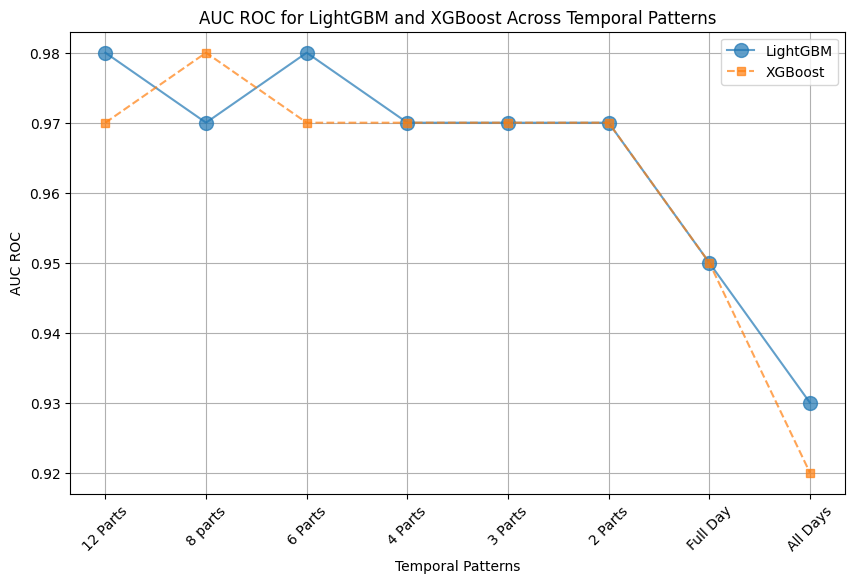}
\caption{AUC-ROC values for LighGBM and XGboost models for all 8 temporal patterns }
    \label{fig:first-label}
 \end{figure}

\vspace{5pt}
When all the days of observation were considered together, the AUC-ROC decreased to 0.93. These results highlight the importance of accounting for circadian rhythms and daily patterns in achieving accurate schizophrenia classification. 

\vspace{5pt}
Schizophrenia symptom expression exhibits temporal patterns. Some individuals with Schizophrenia may experience symptoms late at night or early in the morning, causing disruptions in sleep and wakefulness \cite{b27},\cite{b28}. They may exhibit increased motor activity during periods of agitation or psychosis, which occurs more prominently at certain times \cite{b29}. On the other hand, reduced motor activity during specific temporal segments might be associated with depressive symptoms or social withdrawal \cite{b30}. Understanding these temporal patterns provides valuable insights into the condition.

\subsection{Diminishing returns for Fine Granular Segmentation}

Fine granular segmentation involves breaking down a given time frame into precise intervals. This approach captures every possible variation in motor activity patterns throughout the different parts of the day, assuming that finer temporal divisions would lead to better classification results. However, our observations reveal that fine granular segmentation may not significantly improve classification results, particularly for Schizophrenia. This is evident in the fact that when we divided the data into morning, afternoon, evening, and night patterns, we achieved similar classification results to when we employed a simpler 2-part day and night segmentation.

\vspace{5pt}
Schizophrenia symptom expression exhibits temporal patterns but may not be as fine-grained as dividing the day into 4/6/8/12 parts of the day. Instead, these patterns operate in broader time segments, such as morning or night. Furthermore, fine granular segmentation can introduce noise and complexity into the data without necessarily yielding more meaningful insights. We can achieve accurate schizophrenia classification by focusing on practical, interpretable broader temporal patterns like day and night. This simplification enhances the practicality of our classification approach, making it more accessible for real-world applications.


\section{\textbf{Conclusion and future scope}}

Our study investigates the extent of temporal segmentation required for distinguishing schizophrenic patients from the control group. It becomes evident that finer-grained segmentation does not enhance the AUC-ROC of the classification, as broader day and night divisions yield comparable results to the 3, 4, 6, 8 or 12- part divisions of the day. The statistical features employed are robust enough to capture critical variations during the broader periods, and the LightGBM model outperforms other machine learning models. This simplification streamlines the segmentation process into two parts, speeding up early diagnosis and close monitoring of schizophrenia patients by only focusing on day/night segmentation and not wasting time and efforts on further temporal segmentation. This new study regarding temporal segmentation aims to assist practitioners in creating tools or applications that aid in diagnosis and treatment planning based on temporal motor activity data. Future research could explore the determination of optimal temporal segmentation patterns for individualized diagnosis or developing interpretative models that offer insights into the relationship between specific temporal patterns and symptom expression in Schizophrenia. Using findings from this research, personalized wearable devices can be created to optimize efficiency by focusing on capturing and analyzing motor activity data during specific day and night intervals, enabling effective identification of patterns in sleep-wake cycles, all while conserving energy and streamlining data processing. Incorporating multi modal data into the devices like heart rate along with actigraphy could provide more holistic view of mental health conditions.

\section*{\textbf{Acknowledgment}}

We would like to sincerely thank the creators of the 'Psykose' dataset. All the details are available here- https://datasets.simula.no/psykose/. RA would like to thank DBT NNP (BT/PR40236/BTIS/137/51/2022) for funding.

\end{document}